\documentclass[sigplan]{acmart}

\AtBeginDocument{%
  }

\begin{document}

\title{New Solutions on LLM Acceleration, Optimization, and Application}

\author{Yingbing Huang$^1$, Lily Jiaxin Wan$^1$, Hanchen Ye$^1$, Manvi Jha$^1$, Jinghua Wang$^1$, Yuhong Li$^1$, \\ Xiaofan Zhang$^2$, Deming Chen$^1$}

\affiliation{%
\textit{\{yh21, wan25, hanchen8, manvij2, jinghua3, leeyh, dchen\}@illinois.edu, xiaofanz@google.com} \\
$^1$University of Illinois Urbana-Champaign, $^2$Google
\country{}
}


\begin{abstract}
Large Language Models (LLMs) have become extremely potent instruments with exceptional capacities for comprehending and producing human-like text in a wide range of applications. However, the increasing size and complexity of LLMs present significant challenges in both training and deployment, leading to substantial computational and storage costs as well as heightened energy consumption. In this paper, we provide a review of recent advancements and research directions aimed at addressing these challenges and enhancing the efficiency of LLM-based systems. We begin by discussing algorithm-level acceleration techniques focused on optimizing LLM inference speed and resource utilization. We also explore LLM-hardware co-design strategies with a vision to improve system efficiency by tailoring hardware architectures to LLM requirements. Further, we delve into LLM-to-accelerator compilation approaches, which involve customizing hardware accelerators for efficient LLM deployment. Finally, as a case study to leverage LLMs for assisting circuit design, we examine LLM-aided design methodologies for an important task: High-Level Synthesis (HLS) functional verification, by creating a new dataset that contains a large number of buggy and bug-free codes, which can be essential for training LLMs to specialize on HLS verification and debugging. For each aspect mentioned above, we begin with a detailed background study, followed by the presentation of several novel solutions proposed to overcome specific challenges. We then outline future research directions to drive further advancements.
Through these efforts, we aim to pave the way for more efficient and scalable deployment of LLMs across
a diverse range of applications. 
\end{abstract}

\maketitle

\pagestyle{empty} 

\section{Introduction}
In recent years, Large Language Models (LLMs) have emerged as powerful tools across various domains, revolutionizing natural language processing, information retrieval, LLM-aided design, and others. The ability of LLMs to understand, generate, and manipulate human language has propelled them to the forefront of research and applications in various industries. These models, trained on vast amounts of text data, demonstrate unparalleled proficiency in tasks such as text generation, translation, summarization, sentiment analysis, and more \cite{geminiteam2024gemini}\cite{LLaMA3}. Additionally, there are ongoing efforts to train LLMs with groundbreaking multimodal capabilities, encompassing both visual and speech understanding \cite{xu2023training, xu2023conformer, lei2023acoustic, lei2024personalization}. They have been successfully applied in various applications, including virtual assistants, content generation, question-answering systems, and recommendation systems. The versatility and effectiveness of LLMs have made them indispensable tools in various industries, driving innovation and accelerating progress in artificial intelligence. 

In domains such as LLM-aided design, large language models have been utilized for a variety of tasks, including high-level synthesis, hardware description generation, and functional verification, significantly streamlining the design process and reducing time-to-market for hardware designs \cite{MRHMisuDafnyFSE24}. For instance, ChipNeMo \cite{liu2023chipnemo} enhances LLaMA2 with domain-specific optimizations for more efficient hardware design. AutoChip \cite{thakur2023autochip} focuses on automating HDL generation using feedback from LLMs, thereby improving the iterative design process. ChatEDA \cite{10485372} leverages LLMs to create an autonomous agent for EDA, while VeriGen \cite{thakur2023verigen} specializes in generating Verilog code. Additionally, DIVAS \cite{paria2023divas} provides an end-to-end framework for SoC security analysis, and another approach utilizes LLMs to fix hardware security bugs \cite{ahmad2023fixing}. Moreover, large language models are also being utilized for automated code generation for information technology tasks in YAML, further showcasing their versatility and efficiency \cite{10247987}. 

However, the widespread adoption of these models has been hindered by their demanding computational requirements, which often result in slow response times and high costs for hardware and energy. Addressing these challenges is crucial to fully harnessing the potential of LLMs and unlocking their benefits in real-world applications. To address these challenges, this research paper explores a comprehensive approach to optimize LLMs at algorithm, hardware, compiler, and design-automation levels, aiming to enhance their efficiency and performance across diverse applications.


Previous works explore various algorithmic strategies aimed at decreasing the inference latency of LLMs. 
We begin by examining various methods for optimizing parameter utilization in large language models. These methods include techniques such as early exiting and layer skipping \cite{elbayad2019depth}, which help reduce computational overhead, as well as contextual sparsity, which dynamically prunes irrelevant parameters during inference \cite{liu2023deja}. Additionally, previous works explore a Mixture of Experts (MoE) approach, which distribute computation across multiple sub-models to enhance efficiency and scalability \cite{du2022glam}.
We then delve into optimization techniques for Key-Value (KV) cache, which is crucial for complex tasks like chain-of-thought reasoning and information retrieval. Additionally, we discuss advancements in parallel decoding \cite{leviathan2023fast}, including techniques for aligning small and large model predictions, multi-token predictions, and parallel processing capabilities. Building upon this background study, we propose two novel approaches: a parallel decoding framework called Medusa \cite{cai2024medusaICML}, which employs multiple decoding heads coupled with an optimized tree-based decoding strategy, and SnapKV, a method for effectively reducing KV cache size \cite{li2024snapkv}. Our experimental results demonstrate significant speedups in inference time without compromising generation quality, along with improved memory efficiency. Finally, we outline future directions for tailored algorithmic optimization, advancements in KV compression, and tackling the computational load from speculative decoding, aiming to boost LLM efficiency and effectiveness in practical applications.

LLM-hardware co-design aims to customize hardware architectures to meet the demands of LLMs while providing insights to optimize LLM architectures \cite{hao2018deep}. 
Previously, we proposed an LLM-hardware co-design framework called AutoDistill \cite{zhang2022autodistill}, which integrates model compression, transformer architecture exploration, and multi-objective optimization to produce student models with lower inference latency and smaller sizes while maintaining high accuracy. Moreover, a pruning-aware quantization strategy that combines two effective LLM compression methods, pruning, and quantization, to optimize LLM architectures for hardware efficiency has been proposed \cite{wan2024software}. Furthermore, we explore the potential of reconfigurable and heterogeneous hardware for LLMs, aiming to dynamically adjust hardware architectures to accommodate the latest LLM advancements and model compression methods, thereby enhancing both model quality and hardware efficiency.

The demand for efficient hardware accelerators for deep neural networks has led to a new direction of using High-Level Synthesis (HLS) frameworks \cite{xpilot} \cite{HLS-TRETS} to quickly translate model architectures into hardware implementations. However, exploring the vast design space effectively to achieve optimal solutions remains a significant challenge. 
We summarize two novel compilation frameworks published previously by us: ScaleHLS \cite{hpca2022scalehls, jun2023autoscaledse}  and HIDA \cite{ye2023hida}. ScaleHLS leverages the MLIR infrastructure~\cite{lattner2021mlir} for scalable High-Level Synthesis, optimizing hardware designs with a Design Space Exploration engine by performing multi-level transformations. As far as we know, ScaleHLS was the first flow that could take a PyTorch model and transform it into synthesizable C code that can then be translated into RTL code for hardware implementation. HIDA, built on top of the ScaleHLS framework, automates the conversion of algorithmic hardware descriptions into efficient dataflow architectures, also directly generating HLS accelerators from PyTorch models.
Looking forward, we discuss future directions, including spatial architecture exploration, runtime reconfiguration for scalability, and heterogeneous computing solutions, to further enhance the efficiency and scalability of hardware accelerators for LLMs. Through these advancements, we aim to address the computational and memory challenges associated with LLM acceleration, ultimately improving the performance and energy efficiency of hardware implementations.

There has recently been a growing interest in leveraging LLMs to enhance Electronic Design Automation (EDA) processes, offering significant potential for improving design productivity, code generation, and verification \cite{10485372}. Existing research in this domain encompasses various applications, including assistant chatbots for design workflow enhancement, Verilog and script generation, and Hardware Description Language (HDL) verification and analysis. Despite these advancements, several challenges persist, notably the scarcity of high-quality, domain-specific datasets and the need for more specialized LLMs tailored to grasp the intricacies of electronic design languages and processes. As a case study to leverage LLMs for assisting circuit design, we focus on an important task: High-Level Synthesis (HLS) functional verification. We pursue this task through the construction of the Chrysalis dataset, an extensive collection of HLS designs embedded with diverse sets of realistic bugs, and the development of an HLS-specific debugging assistant. A debugging assistant can be built by training an LLM fine-tuned on the Chrysalis dataset which aims to significantly expedite the verification process, enhance productivity, and reduce time-to-market for hardware designs. Additionally, we outline future research directions, including LLM-aided formal verification and the integration of LLMs into multi-agent systems for hardware/software design automation, offering a transformative approach to streamlining the design, verification, and debugging processes in EDA. 

In the rest of the paper, Section 2 delves into algorithm-level acceleration for LLMs, while Section 3 provides an overview of hardware co-design tailored for LLMs. Section 4 focuses on the compiler for mapping LLMs to accelerators, and Section 5 explores LLM aided designs. Finally, Section 6 presents the conclusion of the study.





\section{LLM Algorithm-Level Acceleration}

\subsection{Background Study}
LLMs excel in tasks such as coding copilot, question answering, and summarization. However, their autoregressive nature, where each token depends on all previous ones, makes decoding memory-intensive and hard to parallelize. This results in significant latency due to the massive size of LLM parameters, impacting applications requiring fast responses like chatbots. Addressing the challenge of reducing inference latency in LLMs is becoming increasingly critical.
This section primarily explores previous methods aimed at decreasing the inference latency of LLMs from an algorithmic standpoint, which could facilitate straightforward implementation and integration across various platforms.


\subsubsection{Efficient Parameter Utilization}
The early study \cite{simoulin2021many} shows only a necessary subset of parameters is used per input token to reduce language model inference latency while preserving accuracy.
The concepts of early exiting and layer skipping~\cite{elbayad2019depth, schuster2022confident} in decoder architectures, allow for an efficient generation of sequences by potentially exiting the decoding process early based on certain criteria, thus saving on computational resources while maintaining output quality. 
On another perspective, contextual sparsity, as investigated by Liu et al.~\cite{liu2023deja}, leverages the insight that a significant portion of model weights can be dynamically omitted without affecting performance, capitalizing on the variability of importance across different weights for different inputs.
Lastly, the Mixture of Experts (MoE)~\cite{du2022glam, zoph2022designing, jiang2024mixtral} approach decouples model size from computational demands, enabling significant scaling of model capacity with minimal impact on inference efficiency, offering a pathway to enhancing model performance without proportional increases in computational burden. 

\subsubsection{KV Cache Optimization}
KV Cache in LLMs stores previously computed attention values and keys. This caching proves particularly effective for complex tasks like chain-of-thought~\cite{wei2022chain} reasoning or information retrieval~\cite{lewis2020retrieval}.
However, it introduces overheads like setup time, extra memory for cache storage, and the complexity of managing cache validity when the sequence length or batch size increases. 
Several strategies have been developed to enhance KV Cache efficiency. One key approach is through advanced KV Cache management techniques. For instance, the VLLM~\cite{kwon2023efficient} introduces PagedAttention which stores keys and values in segmented memory blocks, allowing for more efficient retrieval during attention calculations. Additionally, solutions like 
Hydragen~\cite{juravsky2024hydragen} employ a Shared-prefix KV Cache strategy, greatly improving cache reuse rates by leveraging common sequences. Another significant advancement is the use of KV Cache compression~\cite{zhang2023h2o}, which implements eviction policies to selectively retain tokens in the cache, guided by a scoring function based on cumulative attention.

\subsubsection{Parallel Decoding}

Parallel decoding presents a unique approach by executing multiple decoding steps simultaneously to reduce the overall steps needed, diverging from traditional methods. It typically involves a smaller "draft" model predicting several upcoming words, which are then collectively assessed by the main LLM. This technique is specifically adapted for LLM efficiency. 
Recent advancements in parallel decoding for LLMs include techniques by Leviathan et al.~\cite{leviathan2023fast} and Chen et al.~\cite{chen2023accelerating}, which introduce a resampling strategy to align small and large model predictions with the LLM's distribution, ensuring output consistency. 
Stern et al.~\cite{stern2018blockwise} explore multi-token predictions from a single forward pass using a linear projection layer and a tree-based decoding structure to improve decoded sequence acceptance. Additionally, Santilli et al.~\cite{santilli2023accelerating} and Fu et al.~\cite{fu2024break} adapt Jacobi and Gaussian-Seidel algorithms~\cite{ortega2000iterative} for parallelizing decoding, incorporating n-gram reuse and attention masks to enhance LLM efficiency.

\begin{figure}[t]
    \centering
    \includegraphics[width=\linewidth]
    {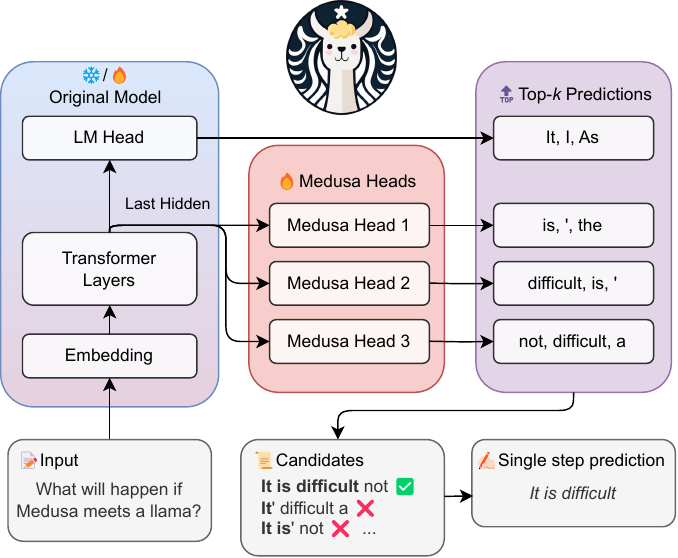}
    \caption{The proposed parallel decoding framework Medusa. During inference, each head generates multiple top predictions for its designated position. These predictions are assembled into candidates processed in parallel using a \emph{tree-based attention} mechanism. Then the framework verifies the candidates and accepts a continuation~\cite{cai2024medusa}. 
    }
    \label{fig: medusa}
\end{figure}

\subsection{Proposed Works}\label{LLM-acceleration}

LLM inference is
predominantly memory-bandwidth-bound with the main latency bottleneck stemming
from accelerators’ memory bandwidth rather than
arithmetic computations. This bottleneck is inherent to
the sequential nature of auto-regressive decoding, where
each forward pass requires transferring the complete model
parameters from High-Bandwidth Memory (HBM) to the
accelerator’s cache. This process, which generates only a
single token, underutilizes the arithmetic computation potential
of modern accelerators, leading to inefficiency. In our proposed work~\cite{cai2024medusa}, named as Medusa, we revisit the concept of parallel decoding with a new perspective, noting that current research primarily aims to boost generation speed through draft models. Yet, obtaining an appropriate draft model either
from scratch or from distillation is non-trivial. Also, hosting dual-sized models on a server presents challenges, and it’s even harder to integrate the
draft model into a distributed system. To tackle this, we present a novel approach (shown in Fig.~\ref{fig: medusa}) using multiple decoding heads as the adapter for prediction, coupled with an optimized tree-based decoding strategy, enhancing the efficiency of the method. Our proposed technique does not need a separate draft model and 
allows for seamless integration into existing LLM systems. Our experiments demonstrate that limited-resource fine-tuning can achieve over 2.2$\times$ speedup without compromising generation quality, while full fine-tuning further improves the speedup to 2.3-2.8$\times$ on models with various sizes. Furthermore, parallel decoding improves resource utilization due to increased matrix operations for multi-token validation per step. By employing an optimized tree-based attention mechanism, we strive to minimize the overhead introduced by parallel decoding.
 Our focus on optimization of both fine-tuning and inference with the decoding adapter in the context of speculative decoding presents a novel direction for enhancing LLM performance.
 
\begin{figure}[ht]
    \centering
    \includegraphics[width=\linewidth]
    {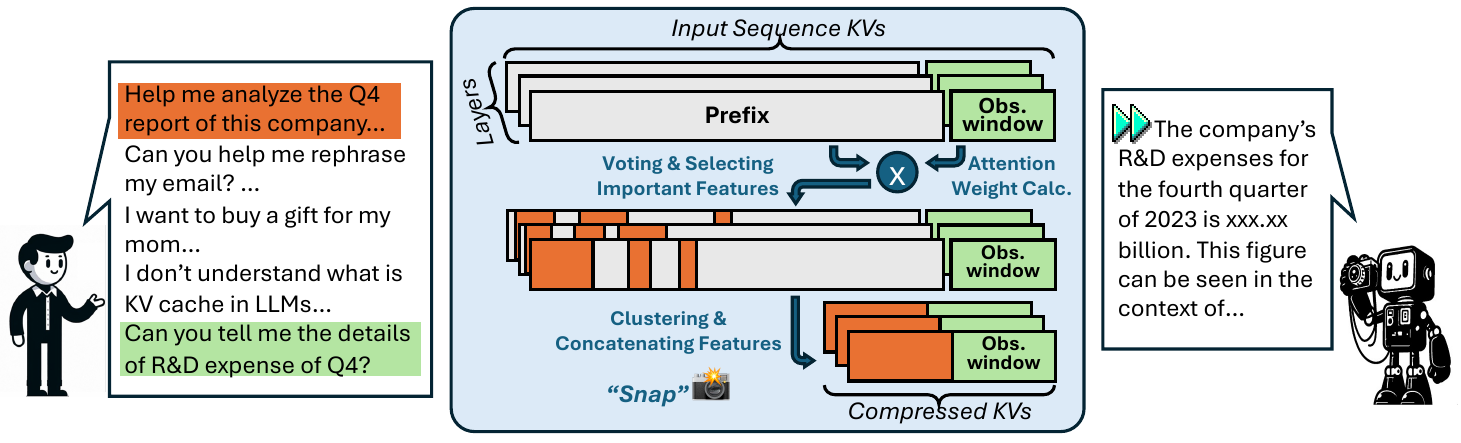}
    \caption{The graph shows the simplified workflow of SnapKV, where the orange area represents the group of positions per head clustered and selected by SnapKV.
    }
    \label{fig: snapkv}
\end{figure}

Furthermore, our method, SnapKV \cite{li2024snapkv}, effectively reduces KV cache size, addressing the computational and memory bottlenecks in scenarios involving long sequence inputs. Our findings demonstrate the consistent attention allocation patterns of important features in prompts used throughout the generation process, independent of prompt formats. This observation highlights the potential on KV cache compression for long sequence input, which could reduce the computational and memory overhead in attention calculation during generation steps. Leveraging this insight, our innovative approach intelligently identifies these attention allocation patterns by using the window of features at the end of long sequence input, as shown in Fig. ~\ref{fig: snapkv}, and compresses the KV cache  accordingly. This proposed work achieves consistent decoding speeds, significantly enhancing generation speed by 3.6x and improving memory efficiency by 8.2x compared to the baseline, when processing inputs of 16k tokens.

\subsection{Future Directions}

\subsubsection{Enhanced Versatility in Parallel Decoding}
With the growth in the size of LLMs and their deployment across both cloud and edge devices, accurately predicting the performance of parallel decoding models has become increasingly complex. A universal solution for LLM performance optimization remains elusive. Without accurate prediction before training and deployment, there is a risk of wasting computational resources and failing to meet performance targets. We focus on modeling and predicting various scenarios, utilizing an analytic model to achieve precise performance estimations. Our objective is to create predictive frameworks that aid in selecting optimal parallel decoding algorithms and their hyperparameters tailored to model size, task complexity, and performance goals. This aims to enhance efficiency, adaptability, and overall model performance.

\subsubsection{Combining KV Compression and Parallel Decoding}
Leveraging KV compression, we see opportunities for notable improvements in tasks with large input prompts, like summarization and multi-round chats, where precise prompt compression will be crucial for maintaining retrieval accuracy and understanding. In long-context scenarios, directly processing the entire prompt and performing inference with parallel decoding introduces significant inference overhead due to the increased computational complexity and memory requirements. To address this, we explore effective attention mechanisms such as Group Query Attention~\cite{ainslie2023gqa} and techniques like quantizing the KV cache to reduce computational load. These refinements are intended to boost LLM efficiency and effectiveness in practical uses while maintaining the generation quality.

\section{LLM-Hardware co-design}

\subsection{Background Study}
The exceptional capabilities of LLMs are countered by their significant memory and computational overheads. Addressing these, LLM-hardware co-design, inspired by the DNN-accelerator co-design methodology \cite{hao2018deep} \cite{hao2019fpga}, customizes hardware to meet these demands and provides insights to optimize LLM architectures. Specialized accelerators, like GPUs, TPUs, and FPGAs, enhance parallel processing and memory capacity and provide efficient LLM execution. At the same time, software strategies, such as model distillation, pruning, and quantization, can effectively reduce LLM size and complexity, making them adaptable to hardware constraints.

We have seen customized hardware accelerators are built for LLM workloads. For example, Tensor Processing Units (TPUs) \cite{jouppi2023tpu} are designed to efficiently handle matrix operations which are fundamental for LLM attention and linear layers. 
These accelerator designs enhance LLM support by integrating High Bandwidth Memory (HBM), reconfigurable high-speed interconnects, and multi-type parallel computation support, offering cost-effective LLM training and serving solutions. 
Beyond ASICs, FPGA-based accelerators are being actively investigated for their potential to provide more flexible and faster turnaround solutions. For example, DFX \cite{hong2022dfx} utilizes model parallelism and enables rapid concurrent execution of transformer-based workloads, while FlightLLM \cite{zeng2024flightllm} introduces a configurable compute unit and a LLM mapping flow to support LLM inference.


For LLM designs, researchers have investigated hardware-aware model compression technologies to optimize LLM architecture.
FlashAttention \cite{dao2022flashattention} reduces the number of High Bandwidth Memory (HBM) accesses by using tiling techniques in attention computations and extends it to block-sparse attention. PagedAttention \cite{kwon2023efficient} divides the KV cache into blocks and manages blocks as pages in OS’s virtual memory, reducing the internal and external fragmentation and thus increasing the efficiency within a single request.
In addition, model distillation, pruning, and quantization have proven to improve hardware efficiency for LLM deployment. MLFS \cite{kundu2024efficiently} freezes a base model and stores many small low-rank adapter matrices, which maintains high quality encoder models on edge applications and reduces training time.
LLM.int8~\cite{dettmers2022llm} develops a two-part vector-wise quantization procedure and a new mixed-precision decomposition scheme, enabling models like OPT-175B on a single server with consumer GPUs. SmoothQuant~\cite{xiao2023smoothquant} uses a per-channel smoothing factor to handle outliers in activations and achieves up to 1.56x speedup and 2x memory reduction. 
ViTCoD \cite{you2023vitcod} prunes attention maps to either dense or sparse patterns and designs an accelerator that coordinates between these two workloads to boost hardware utilization while integrating on-chip encoder and decoder engines. 


\subsection{Proposed Works}

\begin{figure}[t]
    \centering
    \includegraphics[width=\linewidth]
    {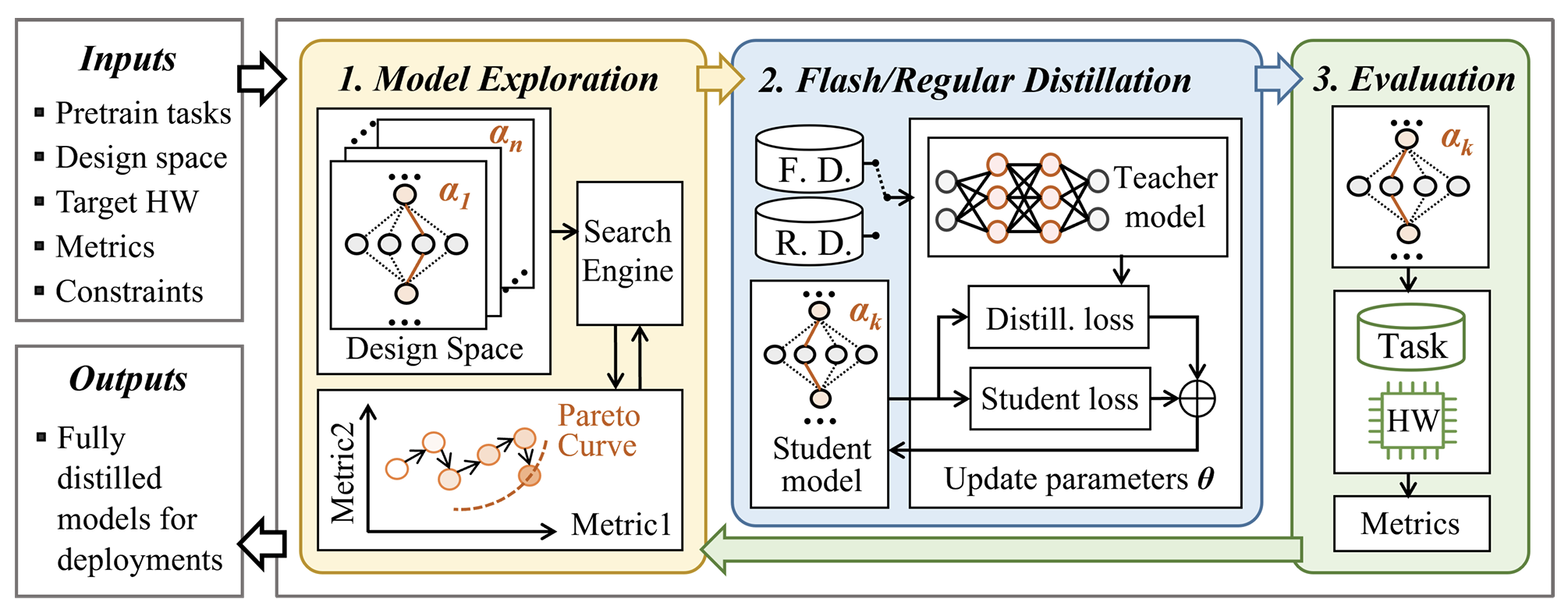}
    \caption{The proposed AutoDistill framework \cite{zhang2022autodistill}. 
    }
    \label{fig: autodistill}
\end{figure}

Following the LLM-hardware co-design method, we propose AutoDistill \cite{zhang2022autodistill}, an end-to-end model distillation framework to deliver hardware-efficient models. As shown in Fig. \ref{fig: autodistill}, AutoDistill introduces a three-stage solution, which includes model architecture exploration, model compression, and model evaluation to deliver efficient models given the target hardware and hardware-software tradeoff requirements. To facilitate the hardware/software co-design process, these stages are tightly connected and continuously iterated in a quality-performance space. During the evaluation stage, model quality and its hardware performance results are passed back to the model exploration to guide the search engine for finding a better model architecture that could fulfill both software and hardware requirements. Results show that AutoDistill can efficiently produce student models with lower inference latency and smaller sizes, while maintaining high accuracy on multiple downstream tasks, such as SQuAD for question answering and reading comprehension in Table \ref{tab:squad}. 

\begin{figure}[ht]
    \centering
    \includegraphics[width=\linewidth]
    {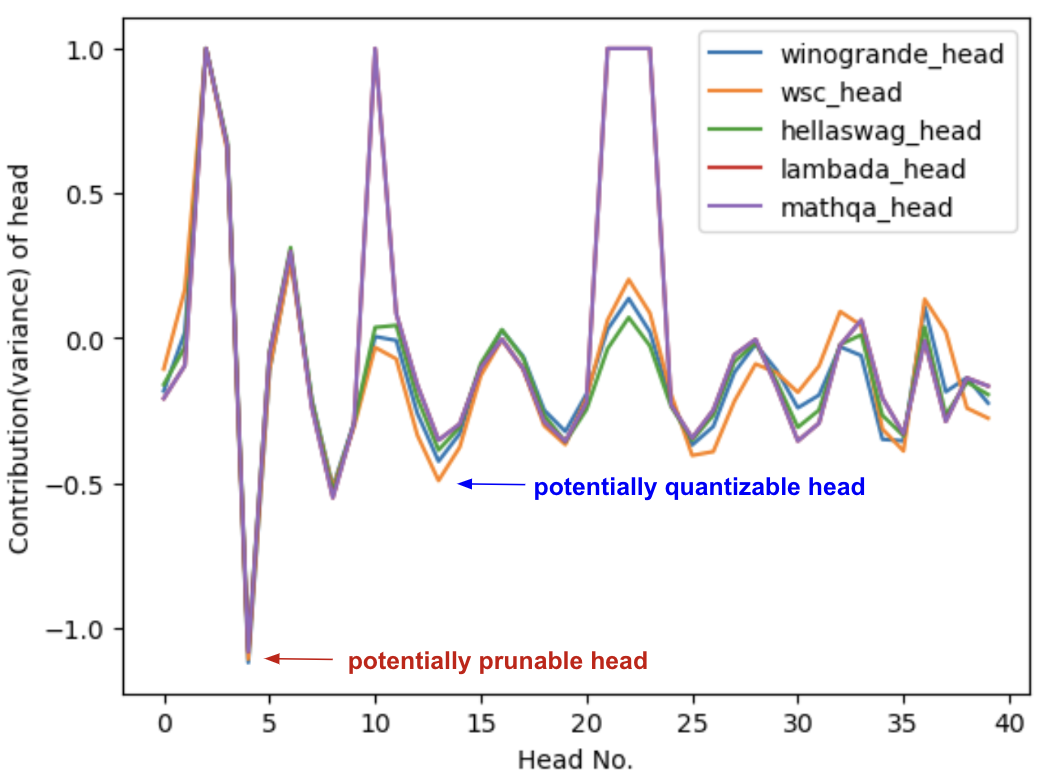}
    \caption{The profiling results on the activity of heads across different datasets by measuring each head's contribution based on its variance over the input sequence. Heads that show low variance are considered inactive, leading to contextual sparsity.
    }
    \label{fig: attn_head}
\end{figure}

\begin{figure}[ht]
    \centering
    \includegraphics[width=\linewidth]
    {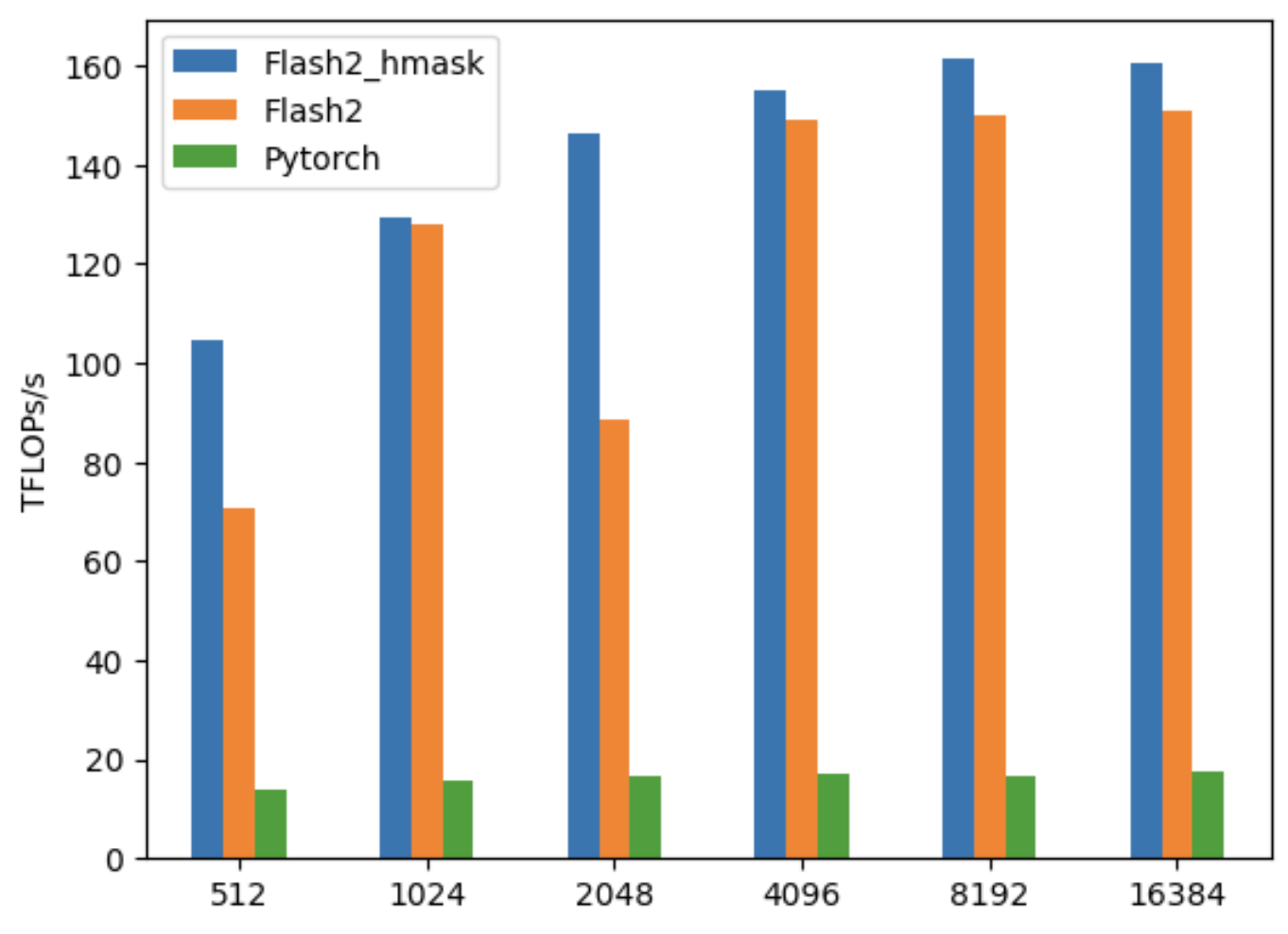}
    \caption{The preliminary result from forward throughput improvement. Flash2\_hmask is the result from the combination of FlashAttention2 \cite{dao2022flashattention} and our
pruning-aware quantization approach \cite{wan2024software}.  
    }
    \label{fig: hmask}
\end{figure}

We also propose a pruning-aware quantization strategy by combining two of the most effective LLM compression methods, pruning and quantization, for LLM-hardware co-design ~\cite{wan2024software}. We observe a similar sparsity distribution pattern of attention heads across various datasets as shown in Fig. \ref{fig: attn_head}, which could potentially be used to smartly choose either completely pruning (0 bit) or different quantization precision (4, 8, 16 bits) for attention heads on individual layers without additional overhead, based on the hardware level objective. Moreover, pruning-aware quantization method could also be combined with other state-of-the-art hardware-aware LLM acceleration frameworks, such as flash attention in Fig. \ref{fig: hmask}, and give higher throughput. 


\begin{table}[t]
\vspace{-8pt}
\caption{The results on SQuADv1.1. Ours-1 and Ours-2 denote two models designed with Autodistill \cite{zhang2022autodistill}.}
\label{tab:squad}
\vskip 0.15in
\begin{center}
\begin{small}
\begin{tabular}{lcccc}
\toprule
  & \# Param  & Latency &  F1  & EM\\
\midrule
BERT$\rm_{BASE}$  & 109 M & - & 88.5 & 80.8\\ \hline
DistilBERT          & 67 M & -        & 85.8 & 77.1 \\
DistilBERT*      & 67 M & -        & 86.9 & 79.1 \\
TinyBERT$\rm_6$*            & 67 M & -        & 87.5 & 79.7 \\
NAS-BERT*          & 60 M & -        & 88.0 & 80.5 \\
NAS-BERT*\dag       & 60 M & -        & 88.4 & 81.2 \\ 
MobileBERT          & 25.3 M          & 0.65 ms  & \textbf{90.0} & \textbf{82.9} \\ 
MobileBERT\ddag     & 25.3 M          & 0.65 ms  & 87.7 & 80.0 \\ \hline

Ours-1  & 22.8 M  &  0.59 ms  & 88.4 & 80.8 \\
Ours-2  & \textbf{20.6 M}  &  \textbf{0.49 ms}  & 88.1 & 80.5 \\
\bottomrule
\end{tabular}
\end{small}
\end{center}
\vskip -0.1in
\end{table}

\subsection{Future Directions}

\subsubsection{System-aware algorithmic optimization}
As LLMs continue to grow in size and complexity, it is becoming increasingly critical to design them with hardware system in mind. It means the model developers should consider the hardware system configurations, including accelerator architecture, compute power, memory capacity, system topology, network bandwidth, model parallelism strategies, in addition to the LLM design parameters. 

This future direction will create multiple optimization opportunities to explore the combined design space consisting of hardware and software configurations. Model quality will not be the only optimization objective while hardware performance and efficiency metrics, such as Queries per second (QPS) and latency, will also be considered and explored as parts of the LLM designs. 
With such an enhanced design space, LLMs can be specifically tailored to the underlying system and hardware architecture and we anticipate further innovations in model size reduction, efficient sharding strategies, optimized data layouts, and other techniques to fully utilize the full potential of target systems.

\subsubsection{Reconfiguratble and heterogeneous hardware for LLMs}
Reconfigurable hardware, such as FPGAs, is a promising solution to address the continuously evolving LLM designs. It offers the ability to adapt to the specific computational patterns of different LLM workloads, which allows a fast development of hardware accelerators to efficiently handle key LLM operations, including matrix multiplication and attention mechanisms. Additionally, it can be combined with heterogeneous hardware to explore new compute paradigms, such as adopting in-memory computing, to address memory-bound operations. This direction enables trade-offs between model quality and hardware efficiency.   

\subsubsection{Co-design for edge LLM applications}
The co-design for edge LLM applications is crucial, given the intricate challenges posed by edge computing's energy and resource limitations. LLM-hardware co-design emerges as a promising solution to these challenges, aiming to harmonize software and hardware to optimize LLM performance on edge devices. Future research will focus on creating tailored architectures and algorithms that efficiently manage computational resources, ensuring that the quality of LLM services remains high. This could involve exploring adaptive power management techniques, optimizing memory usage, and enhancing processing speeds without sacrificing the accuracy or responsiveness of LLM applications.


\section{LLM-to-Accelerator Compiler}

\subsection{Background Study}\label{LtoA}

High-Level Synthesis (HLS) \cite{xpilot}\cite{HLS-TRETS} is vital for rapidly developing efficient, high-density hardware accelerators, enabling quick evaluation of different algorithmic choices. The challenge of enabling scalable compiler from LLM models to HLS-based accelerators lies in effectively exploring the vast design space, which can lead to sub-optimal solutions if not done well, undermining the productivity benefits of HLS. To tackle this challenge, in this section, we will introduce two compilation frameworks, ScaleHLS and HIDA, which can generate HLS accelerators directly from PyTorch models.

\subsection{Proposed Works}

\begin{figure}[h]
    \centering
    \includegraphics[width=\linewidth]{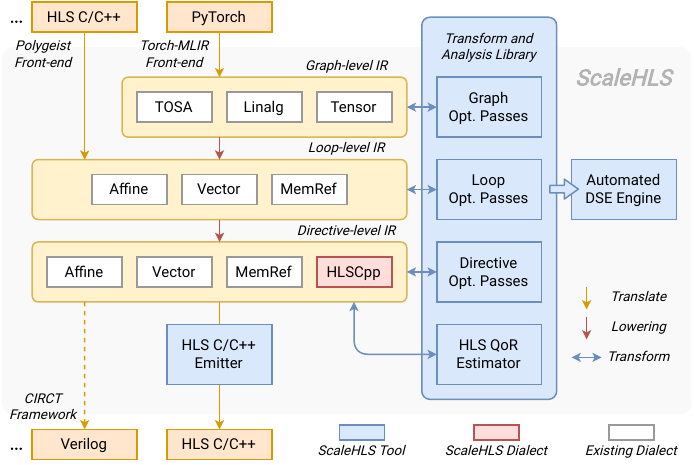}
    \vspace{-15pt}
    \caption{ScaleHLS framework architecture~\cite{hpca2022scalehls}.}
    \label{fig:scalehls_framework}
    \vspace{-5pt}
\end{figure}

\subsubsection{ScaleHLS}
ScaleHLS~\cite{hpca2022scalehls,dac2022scalehls,jun2023autoscaledse,ye2023high} is a scalable HLS framework based on Multi-Level Intermediate Representation (MLIR)~\cite{lattner2021mlir}. Fig.~\ref{fig:scalehls_framework} shows the overall architecture. ScaleHLS supports C/C++ and PyTorch as design entries. Once the inputs are parsed, ScaleHLS supports three levels of IR, \emph{graph}, \emph{loop}, and \emph{directive}, to apply the HLS-oriented optimizations progressively. At the graph and loop level, graph optimizations (e.g., node fusion and coarse-grained pipelining) and loop optimizations can be performed efficiently. At the lowest directive level, HLS-specific optimizations are applied to fine-tune the hardware micro-architecture.

On top of each level of IR, ScaleHLS provides a set of transform passes to optimize HLS designs. By performing each transform pass at the "correct" level of abstraction, ScaleHLS is able to leverage the intrinsic hierarchy of HLS designs and reduce the algorithmic complexity of transforms. Meanwhile, we propose a Design Space Exploration (DSE) engine to automatically optimize the configurable design parameters and search for the Pareto-dominating design points in the latency-resource utilization space. Finally, the optimized IR is emitted as synthesizable HLS C/C++ code. Experimental results show that, comparing to the baseline designs without manual directives insertion and code-rewriting, that are only synthesized by Vitis HLS, ScaleHLS improves the performances with up to 3825.0$\times$ on representative neural network models.

\begin{figure}[h]
    \centering
    \includegraphics[width=\linewidth]{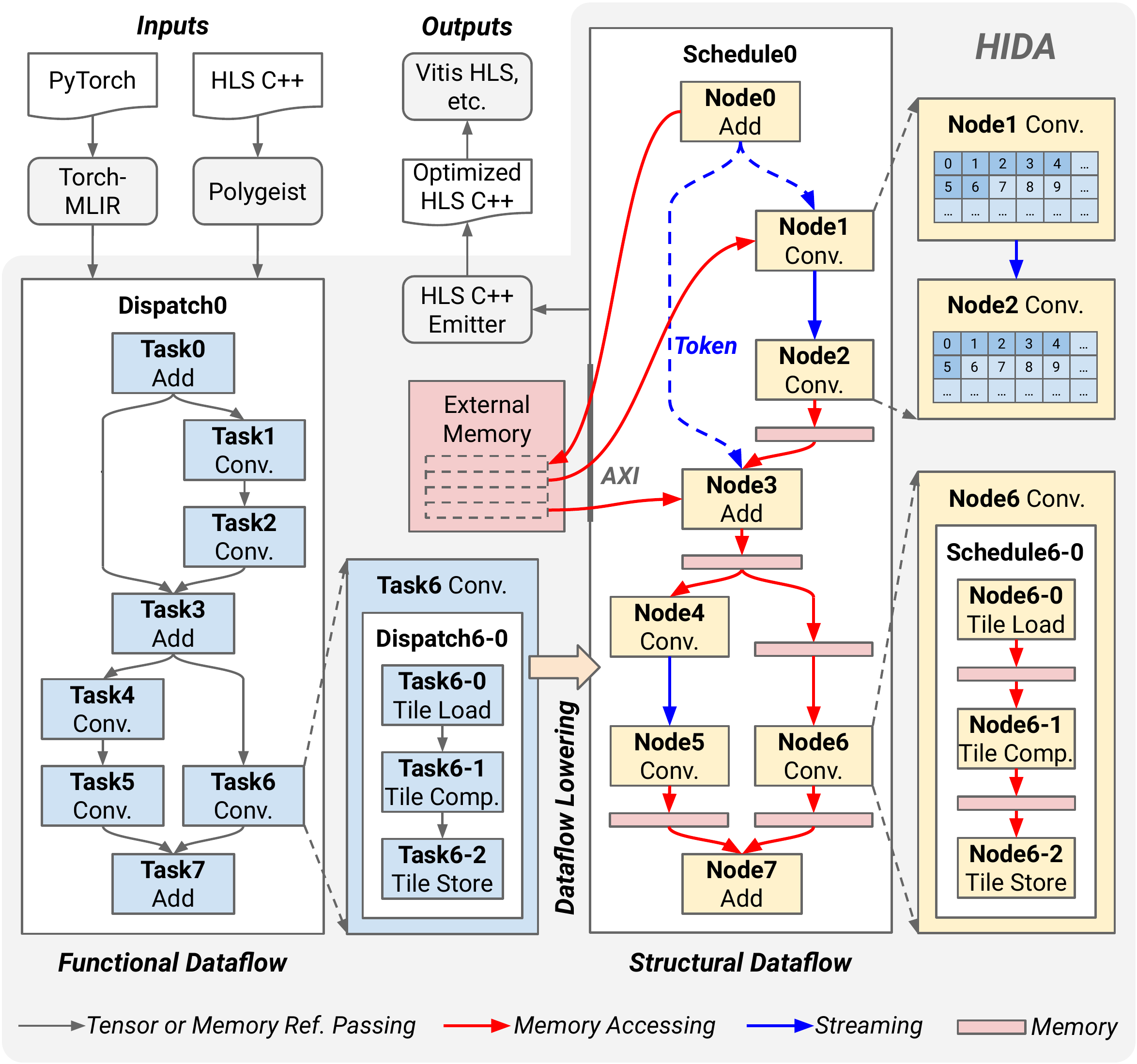}
    \vspace{-15pt}
    \caption{HIDA framework architecture~\cite{ye2023hida}.}
    \label{fig:hida_framework}
    \vspace{-5pt}
\end{figure}

\subsubsection{HIDA}
HIDA~\cite{ye2023hida} is an HLS framework built upon ScaleHLS with \underline{hi}erarchical \underline{da}taflow intermediate representations~(IR) and optimizations, enabling the automated transformation of algorithmic hardware descriptions to efficient dataflow architectures. Fig.~\ref{fig:hida_framework} shows the HIDA's overall architecture. The core of HIDA is its novel dataflow IR, named HIDA-IR, which is designed for modeling dataflow at two distinct abstraction levels: \emph{Functional} and \emph{Structural}. This dual-level approach is critical for capturing the dataflow's characteristics and its multi-level hierarchy, thereby facilitating effective optimizations.

Another important aspect of HIDA is the introduction of HIDA-OPT, a new dataflow optimizer. This optimizer utilizes a pattern-driven task fusion algorithm coupled with an intensity- and connection-aware dataflow parallelization algorithm, which can capture the computation complexity and interconnection topology of the dataflow nodes during the parallelization process. Furthermore, HIDA is designed to be end-to-end and extensible, supporting both PyTorch and C++ inputs. This flexibility empowers users to rapidly experiment with various design parameters and prototype new dataflow architectures, broadening the framework's applicability and ease of use. Despite being fully automated and able to handle various applications, HIDA achieved throughputs that were 8.54$\times$ and 1.29$\times$ higher than those of ScaleHLS and RTL-based neural network accelerators~\cite{zhang2018dnnbuilder}, respectively.

\subsection{Future Directions}

\subsubsection{Spatial Architecture}
Due to the substantial volume of parameters and intermediate computations involved in LLMs, the bottleneck in hardware acceleration frequently resides in the external memory bandwidth.
Contrary to the Von Neumann architecture, which consistently battles the "memory wall," spatial architectures can leverage on-chip communication among tasks to minimize frequent external memory accesses. By overlapping the execution of distinct LLM layers and buffering only a subset of intermediate results on-chip, spatial architectures can markedly decrease on-chip memory requirements and overall latency. This approach presents a compelling solution for LLM inference. Nonetheless, the automatic generation of spatial architectures remains challenging, opening vast avenues for innovation in compilation and architecture design.

\subsubsection{Runtime Reconfiguration}
To achieve spatial parallelization, tasks must be instantiated simultaneously on-chip. However, due to constrained computational and on-chip memory resources, it is infeasible to simultaneously map all layers of emerging LLMs on-chip, which significantly limits the scalability of spatial architectures. Consequently, runtime reconfiguration emerges as a crucial strategy for enabling scalable spatial solutions.
The main challenge lies in automating the balance between spatial and sequential execution — that is, addressing the scheduling problem — to optimize the performance-energy trade-off.

\subsubsection{Heterogeneous Computation}
Accelerating LLMs presents a unique challenge due to their dual nature, being both computation-bound and memory-bound. The prefill phase of LLMs is dominated by General Matrix Multiply (GEMM) operators, making it computation-intensive. In contrast, the generation phase is dominated by General Matrix-Vector (GEMV) operations, demanding substantial memory bandwidth to keep the computation units engaged (refer to Section \ref{LLM-acceleration}). This dual nature of LLMs unveils significant opportunities for heterogeneous computing solutions, where compilers assume an important role in the code generation for heterogeneous platforms and facilitating efficient communication between them.

\subsubsection{Advanced HIDA} Although the HIDA framework can conduct effective dataflow-aware design space exploration, optimizing streaming buffers in LLM accelerators remains a formidable challenge due to the self-attention mechanism and complex inter-layer connections in LLMs. Enhancements in the HIDA framework could address more complicated stream optimizations to reduce on-chip memory consumption. Additionally, recent works~\cite{chen2024allo, zhuang2023charm} have demonstrated the ability to generate efficient kernel designs through customized scheduling. We propose to integrate these highly-optimized kernels into the HIDA explorer to further improve the efficiency of LLM accelerators. We also propose to enhance the code generation of HIDA to support more hardware platforms with dataflow architecture, such as AMD Versal ACAP~\cite{Versal_ACAP}.

\section{LLM-aided design}
\subsection{Background Study}

The existing related work regarding leveraging LLMs in the field of EDA could be divided into three categories \cite{zhong2023llm4eda}: (1) Assistant Chatbot for Enhanced Design Workflow: ChipNemo \cite{liu2023chipnemo} leverages domain adaptation techniques such as custom tokenizers, domain-adaptive continued pretraining, and Supervised Fine-Tuning (SFT) atop the foundation model LLaMA2 \cite{LLaMA2}. This integration facilitates instant access to knowledge, streamlines the query-response process, and diminishes reliance on conventional search methods and associated delays. 
(2) HDL and Script Generation: LLMs, such as those in AutoChip \cite{thakur2023autochip} and VeriGen \cite{thakur2023verigen}, have shown their effectiveness in generating Verilog codes and EDA tool scripts from natural language instructions; 
(3) HDL Verification and Analysis: RTLFixer \cite{tsai2023rtlfixer} exemplifies this by introducing a framework aimed at correcting Verilog code errors utilizing tools like OpenAI GPT-3.5/GPT-4, supplemented by Retrieval Augmented Generation (RAG) and ReAct prompting techniques. Additional efforts in this area focus on generating SystemVerilog Assertions (SVA) for security purposes \cite{kande2023llm}\cite{meng2023unlocking}\cite{paria2023divas}, illustrating the wide-ranging potential of LLMs in bolstering HDL verification and analysis processes.
CHIRAAG \cite{mali2024chiraag} is proposed to generate SVA assertions from natural language specification based on GPT-4. For those assertions with syntax error or simulation error, LLMs could receive the automatic feedback of log file and then regenerate the SVA for retest.
Orenes-Vera \cite{orenes2023using} proposed an iterative methodology where LLMs, particularly GPT-4, are prompted with refined rules and RTL code to generate SVA, which is then evaluated for correctness and completeness through a series of testbench simulations and revisions. 

Despite the promising developments in LLM-aided design within EDA, several challenges remain: (1) Data Quality and Availability: The efficacy of LLMs in EDA critically hinges on the availability of high-quality, domain-specific datasets for their training and refinement. Unfortunately, the proprietary nature of many electronic designs and the tools used for EDA significantly limit access to such datasets. The bulk of detailed hardware design codes, primarily developed within corporate settings, are not made public. This restriction leads to a scarcity of accessible, high-grade datasets, thus hindering the development and optimization of LLMs specifically engineered for EDA applications; (2) Development of Specialized LLMs: There is a critical need for the development of LLMs that are specifically tailored to grasp the complexities of electronic design languages and processes. The generic models, while useful, often lack the nuanced understanding required to effectively generate, verify, and analyze hardware code and to interact with EDA tools at a level that matches human experts. This necessitates a concerted effort to create more specialized models that can comprehend and manipulate the intricate details of electronic designs with a high degree of accuracy and efficiency.

\subsection{Proposed Works}
One use case of LLM-aided design is to harness LLMs for enhancing the verification and debugging processes for HLS code development. HLS, with its higher level of abstraction, can significantly improve design productivity as explained in Section~\ref{LtoA}. 

\subsubsection{Chrysalis Dataset}
\begin{figure}[t]
    \includegraphics[width=\linewidth]{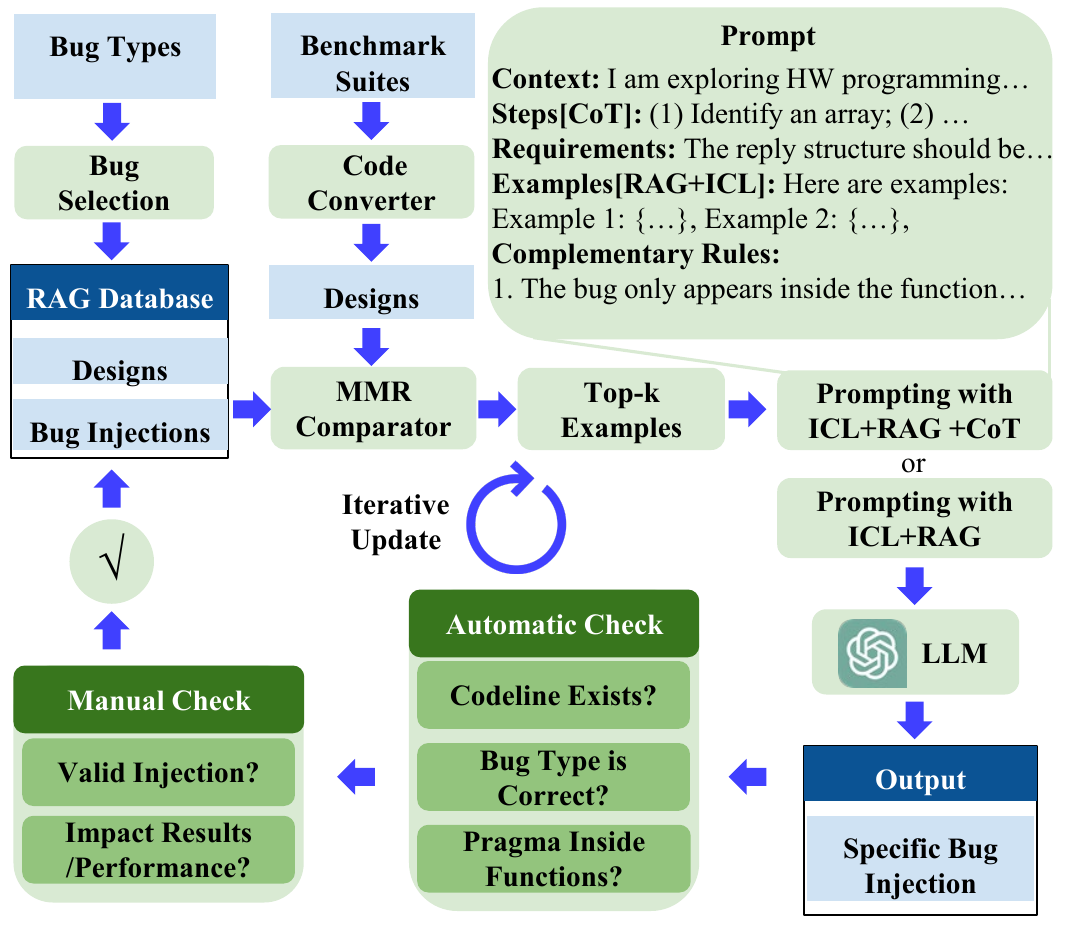}
    \vspace{-15pt}
    \caption{Overview of the \textit{Chrysalis} Dataset Construction and Iterative Upgrade Process: For each check iteration, it involves evaluating the dataset's validity and expanding the RAG dataset accordingly. Through these iterations, the quality of the dataset progressively improves.}
    \label{Chrysalis_Gen}
    \vspace{-5pt}
\end{figure}

The cornerstone of our proposed work is the Chrysalis dataset, an extensive collection of HLS designs embedded with a diverse set of realistic bugs \cite{wan2024software}. This dataset is meticulously curated from a wide range of open-source HLS benchmark suites, featuring over 1,500 designs, with both the version embedded with bugs and the corresponding bug-free version. 
Fig. \ref{Chrysalis_Gen} outlines our methodology for constructing the Chrysalis dataset. We begin by gathering open-source benchmark suites and difficult bug types (bugs also include non-ideal HLS Pragma insertions), which compilers often struggle to identify. These suites are then converted to the function-level designs. Using the Maximal Marginal Relevance (MMR) algorithm, we select the top-k similar designs from the RAG database for bug injection prompts. The prompt generation chooses one strategy based on bug type statistics: one combining In-Context Learning (ICL), Retrieval-augmented Generation (RAG), and Chain-of-Thoughts (CoT); the other using just RAG and ICL. After integration, the prompts are processed by an LLM (GPT-4 in our case) to generate bug (or non-ideal Pragma) injection solutions, which are then validated through both automatic procedures as well as manual checks by hardware engineers. Successful solutions are added to the RAG database to enhance its diversity and volume, improving future solutions. Our evaluations show 88\% validity for all the bugs.
This dataset serves not only as a tool for training our domain-specific LLM, but also as a benchmark for evaluating the model's proficiency in identifying and suggesting fixes for common and complex HLS bugs.

\subsubsection{HLS-specific Debugging Assistant}
\begin{figure}[t]
    \includegraphics[width=\linewidth]{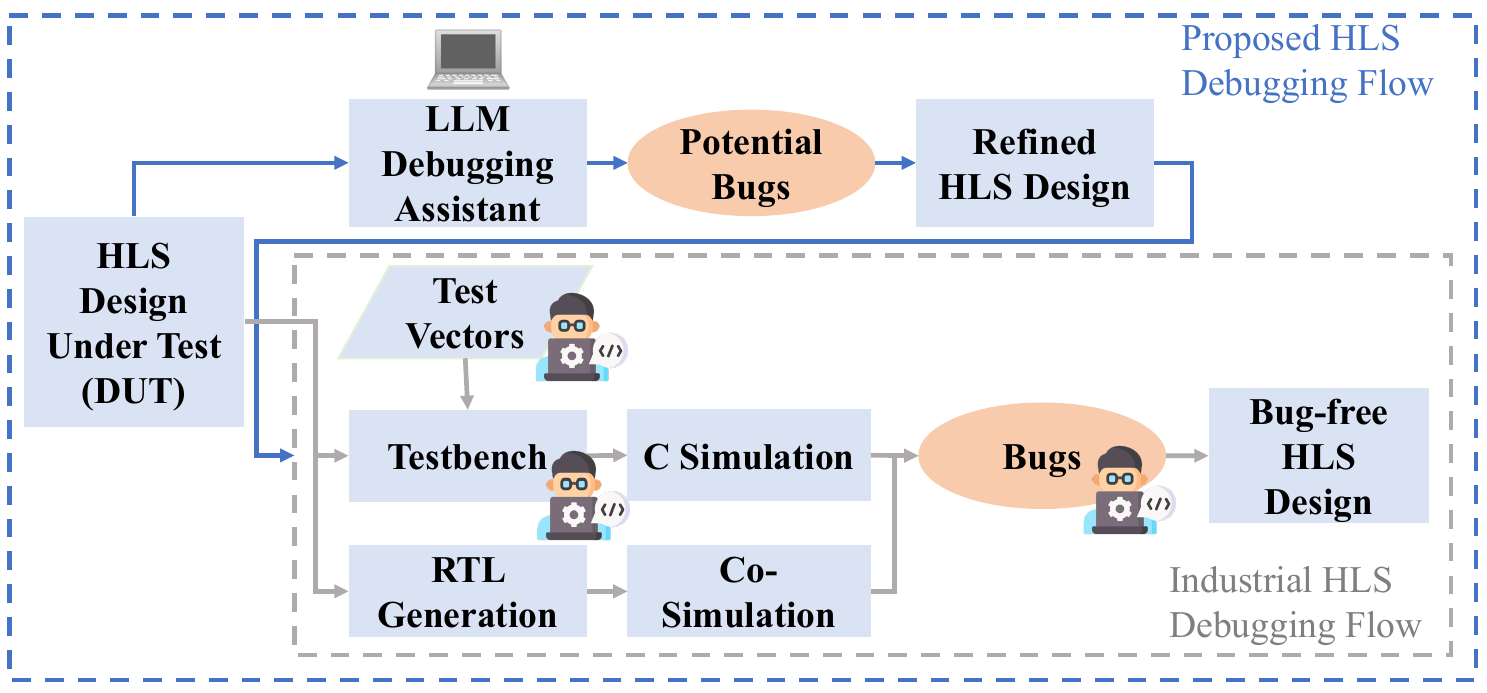}
    \vspace{-15pt}
    \caption{LLM-based HLS Debugging Flows Working Together with Traditional Flows: By incorporating our LLM debugging assistant, the number of bugs requiring verification by test cases can be significantly reduced.}
    \label{flow_comparison}
    \vspace{-5pt}
\end{figure}

Building upon the Chrysalis dataset, our next step involves the creation of an HLS-specific debugging assistant, as Fig. \ref{flow_comparison} shows. Engineers typically design test vectors and create test benches manually, then perform C simulations and co-simulations to analyze and identify potential bugs, which is time-consuming. To improve the efficiency of the debugging process, we proposed a novel flow leveraging the capabilities of LLMs on top of the traditional HLS debugging flow.
This LLM will be fine-tuned to understand the intricacies of HLS code, enabling it to identify errors, suggest corrections, and provide guidance directly within the developers' workflow. The assistant aims to integrate seamlessly with popular development environments, offering real-time support to engineers as they navigate the complexities of HLS designs. By providing context-aware suggestions and corrections, the debugging assistant will significantly expedite the verification process, enhancing productivity and reducing the time-to-market for hardware designs.

The entire methodology could be adapted to RTL debugging as well, starting from the bug injection stage, using open-source LLMs and developing a domain-specific RTL debugger through fine-tuning. To effectively transition to this new application, we must tackle diverse bug types specific to RTL, such as those related to timing constraints, race conditions, and synthesis-simulation mismatches. Particularly, the inherent timing characteristics of RTL designs can lead to more complex bugs, often manifesting as issues in timing analysis that are not present in higher-level abstractions. Given the complexity of RTL code, one ambitious goal is to reduce manual debugging and verification effort by building an advanced and automated RTL verification and debugging flow. This would involve enriching our dataset to include RTL designs (can be generated through HLS working with our Chrysalis dataset) together with test benches and test vectors. A flow similar to Fig. \ref{Chrysalis_Gen} can be developed to assess and improve the validity of such bug injections. Furthermore, seamless integration with EDA tools is crucial to enable real-time analysis and correction within the existing design frameworks.

\subsection{Future Directions}

\subsubsection{LLM-Aided Debugging}
Our research highlights challenges in using LLMs to inject certain HLS bug types, such as operator precedence errors and incorrect data access patterns for interface pragmas. These difficulties stem from sparse code patterns and the complexities of the existing codebase, necessitating further investigation and refined methodologies for effective bug injection. Additionally, as manual verification of bug injections remains necessary in our current flow, creating an automated flow to estimate performance could speed up the identification and resolution of non-ideal Pragma issues, thus enhancing the quality and quantity of the dataset. 
Furthermore, for the HLS-specific debugging assistant, we will employ Low-Rank Adaptation (LoRA) \cite{hu2021lora} for supervised fine-tuning on state-of-the-art open-source LLMs such as LLaMA3 \cite{LLaMA3}, utilizing commercial HLS documentation for design guidelines and rules together with our Chrysalis dataset.

\subsubsection{LLM-Aided Formal Verification}

LLMs can enhance the formal verification process in hardware design by generating precise assertions for the proof of correctness. By integrating these assertions into the formal verification workflow, LLMs can substantially increase hardware design productivity. One promising direction is to explore an iterative process: after the initial proof attempts, the theorem prover's feedback is utilized to refine the LLM's output. This feedback loop enables the LLM to adjust its generated proofs iteratively until the assertions are fully verifiable. Through this dynamic interaction between LLMs and theorem provers, the generation of program proofs becomes both faster and more achievable. This methodology not only speeds up the verification process but also ensures a higher degree of reliability in hardware design verification.

\subsubsection{LLM-Aided Hardware/Software Design Automation}
In the realm of EDA, employing LLM multi-agent systems promises a transformative approach to streamlining the design, verification, and debugging processes. These sophisticated systems autonomously manage various phases of the workflow, seamlessly transitioning from design to verification and debugging. 
By deploying multiple specialized LLM agents — each adept in distinct facets of the design process such as code generation, verification, error detection, and performance optimization — a highly efficient pipeline is crafted. This orchestrated integration allows the agents to collaboratively refine and optimize the design iteratively, leveraging real-time feedback and comprehensive verification results. 
Throughout the process, hardware engineers are only tasked with overseeing the initial specification and periodically reviewing the outputs from the LLMs to ensure that they align with the design intentions and confirm the reliability of the LLMs' outputs.

\section{Conclusion}

In our study, we focused on optimizing LLMs to reduce inference latency and improve efficiency across various applications. We presented a new method, Medusa, to use multiple decoding heads for prediction, coupled with an
optimized tree-based decoding strategy for parallel token processing to speed up the execution of LLMs. We also proposed a novel method, SnapKV, that effectively reduced KV cache size, addressing the computational and memory bottlenecks in scenarios involving long sequence inputs.

We discussed LLM/hardware co-design to integrate both hardware optimization for efficient execution and model architecture exploration for improved system efficiency while maintaining LLM accuracy. HLS frameworks like ScaleHLS and HIDA were explored for accelerating LLMs directly from PyTorch models, envisioning automated generation of spatial architectures and heterogeneous computing solutions.

We also explored the advancements in LLM-aided design for EDA and discussed a novel flow to create the Chrysalis dataset that can be used to train an LLM-based HLS-specific debugging assistant. A similar strategy can be adopted for building an RTL-specific debugging assistant as well. These methods are promising for streamlining the debugging and verification process of the hardware code development. 

For each aspect mentioned above, we also outlined
promising future directions for further research and exploration. 


\begin{acks}
This work is supported in part by the IBM-Illinois Discovery Accelerator Institute, AMD Center of Excellence at UIUC, AMD Heterogeneous Adaptive Compute Cluster (HACC) initiative, NSF 2117997 grant through the A3D3 institute, and Semiconductor Research Corporation (SRC) 2023-CT-3175 grant.
\end{acks}

\bibliographystyle{ACM-Reference-Format}
\bibliography{main}


\end{document}